\documentclass[runningheads]{llncs}

\usepackage{accv}

\usepackage{accvabbrv}
\usepackage{wrapfig,lipsum,booktabs}
\usepackage{graphicx}
\usepackage{booktabs}
\usepackage{amsmath,amsfonts}
\usepackage{array}
\usepackage{float}
\usepackage{multirow}
\usepackage[accsupp]{axessibility}  

\usepackage[pagebackref,breaklinks,colorlinks,citecolor=accvblue]{hyperref}

\usepackage{orcidlink}

\begin{document}

\title{Tamaththul3D: High-Fidelity 3D Saudi Sign Language Avatars
    from Monocular Video}

\titlerunning{Tamaththul3D}

\author{Eyad Alghamdi\inst{1} \and
    Sattam Altuuaim\inst{2} \and
    Obay Ghulam\inst{1} \and
    Abdulrahman Qutah\inst{1} \and
    Yousef Basoodan\inst{1}}

\authorrunning{Alghamdi et al.}


\institute{University of Jeddah, Jeddah, Saudi Arabia \and
    King Abdullah University of Science and Technology, Thuwal, Saudi Arabia}

\maketitle
\begin{center}
    \centering
    \vspace{-3mm}
    \includegraphics[width=1.0\textwidth]{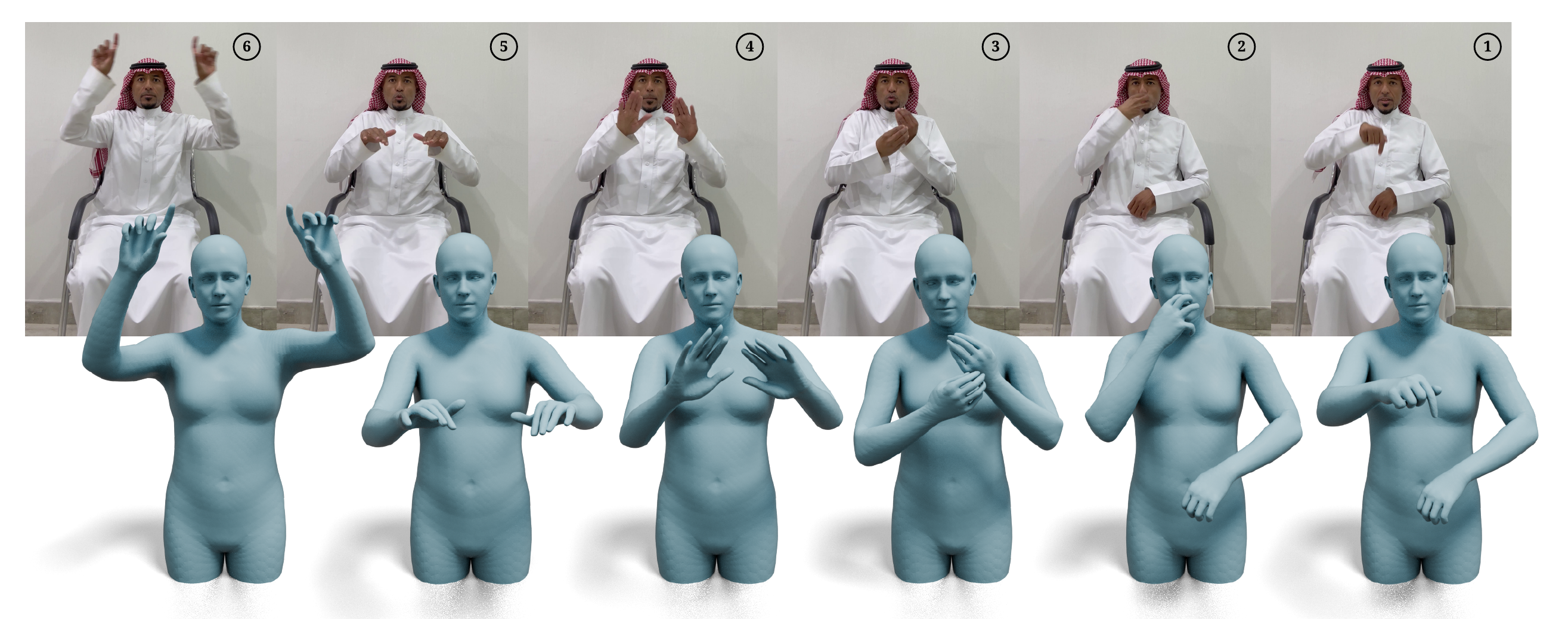}
    \vspace{-6mm}
    \captionof{figure}{\textbf{Tamaththul3D} reconstructs high-fidelity 3D
        sign language avatars from monocular video. \textbf{Top:} a Saudi
        signer performing a Saudi Sign Language sentence. \textbf{Bottom:} the SMPL-X avatars reconstructed
        by Tamaththul3D, recovering accurate hands and body posture despite
        loose traditional attire.}
    \label{fig:teaser}
    \vspace{-2mm}
\end{center}

\begin{abstract}
    Existing 3D sign language avatar reconstruction methods are developed and
    evaluated exclusively on Western sign languages, and no 3D parametric
    annotations exist for any Arabic Sign Language dataset, a gap that blocks
    the development of avatar-based accessibility applications for the Arab
    Deaf community. We release the first SMPL-X parametric annotations for the Ishara-500
    Saudi Sign Language dataset, enabling quantitative evaluation and
    downstream sign language generation for Arabic Sign Language.
    We introduce \textbf{Tamaththul3D}, a reconstruction pipeline that aligns
    hand and body estimates through geometric inverse kinematics on the
    forearm chain followed by 2D-supervised shoulder refinement. The
    closed-form integration is decoupled from the specific choice of
    body and hand estimators: any SMPL-X-compatible body estimator and
    any MANO-compatible hand estimator can be substituted, as we
    demonstrate by swapping each module independently. Tamaththul3D
    achieves up to 32\% lower hand error than prior methods, runs
    $32\times$ faster than the strongest baseline, and generalizes
    across five typologically distinct sign languages without
    dataset-specific adaptation.
\end{abstract}

\section{Introduction}
\label{sec:intro}

Arabic Sign Language (ArSL) and its regional variants, like Saudi Sign
Language (SSL), serve as the primary communication systems for Deaf
communities throughout the Arab world, where sign languages remain the
day-to-day medium of communication. Recent years have seen major
progress in 3D human pose estimation and avatar
reconstruction~\cite{pavlakos2019expressivebodycapture3d,
    cai2024smplerxscalingexpressivehuman, lin2023onestage3dwholebodymesh,
    moon2022accurate3dhandpose}, yet no specialized system exists for
creating detailed 3D signing avatars from ArSL videos with accurate
parametric annotations.

\textbf{The Role of 3D Parametric Annotations.}
High-quality 3D parametric annotations are foundational infrastructure
for sign language avatar research. They enable quantitative evaluation
of reconstruction methods, provide ground-truth supervision for
learning-based approaches, and serve as the motion representation that
downstream sign language generation and translation systems build on
directly~\cite{baltatzis2024neuralsignactorsdiffusion}. For Western sign
languages, datasets such as How2Sign~\cite{duarte2021how2signlargescalemultimodaldataset}
have been enriched with SMPL-X annotations~\cite{baltatzis2024neuralsignactorsdiffusion},
enabling a generation of avatar-based applications. For ArSL, no such
annotations exist. Existing ArSL datasets~\cite{arabsign, karsl} provide
only 2D video, and Ishara-500~\cite{alyami2025isharah}, the largest SSL
dataset, likewise lacks any 3D parametric representation. This absence
blocks the development of avatar-based accessibility applications for the
Arab Deaf community entirely.

\textbf{Limitations of Existing Approaches.}
Existing sign language avatar reconstruction
methods~\cite{Forte23-CVPR-SGNify, baltatzis2024neuralsignactorsdiffusion}
have several shortcomings. SGNify~\cite{Forte23-CVPR-SGNify} uses
linguistic priors but produces visually incorrect hand gestures. Neural
Sign Actors~\cite{baltatzis2024neuralsignactorsdiffusion} focuses on
generating sign language from text and developed a curation process for
SMPL-X datasets using OSX~\cite{lin2023onestage3dwholebodymesh} and
MediaPipe~\cite{lugaresi2019mediapipe}.
DexAvatar~\cite{kundu2025dexavatar} uses sign-language-aware priors for
optimization-based reconstruction but requires 21.60 seconds per frame,
making large-scale annotation impractical. Critically, all existing
methods are developed and evaluated exclusively on Western sign
languages (ASL and German Sign Language), with no consideration for
ArSL's unique characteristics.

\textbf{Unique Challenges of ArSL.}
Sign language reconstruction presents distinct challenges compared to
general human pose estimation~\cite{zheng2021deep}. First, hand
articulation complexity: sign languages convey meaning through precise
finger configurations, palm orientations, and rapid hand motions.
General-purpose pose estimation
methods~\cite{pavlakos2019expressivebodycapture3d,
    lin2023onestage3dwholebodymesh} often fail to capture these intricate
details that render signs incomprehensible. Second, cultural specificity:
ArSL exhibits distinct characteristics in hand shapes, movement patterns,
and signing space that differ from ASL or BSL, and signers frequently
wear culturally specific attire (thobes, abayas, and hijabs) that
obscures body shape cues in models trained on Western data. Third,
multi-modal integration: effective signing requires simultaneous capture
of hand pose, body orientation, and facial expressions, all of which
contribute to semantic meaning.

\textbf{Contributions.}
We introduce \textbf{Tamaththul3D}
(from Arabic
\raisebox{-.3em}{\includegraphics[height=1em]{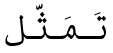}},
meaning ``representation'' or ``likeness''), a reconstruction pipeline
for Arabic Sign Language avatars that generalizes across five
typologically distinct sign languages. Our primary technical
contribution is a geometric forearm alignment method that solves for the
elbow rotation aligning the kinematic chain with WiLoR's global wrist
orientation, combined with 2D-supervised shoulder optimization,
achieving state-of-the-art hand accuracy at $32\times$ the speed of the
strongest baseline.

Our contributions are:
\begin{itemize}
    \item \textbf{First 3D parametric annotations for Arabic Sign Language.}
          We release SMPL-X annotations for the Ishara-500 Saudi Sign
          Language dataset, the first 3D parametric annotations for any
          Arabic Sign Language dataset.
          
    \item \textbf{Geometric hand-body integration.} We solve for the elbow
          rotation that aligns the SMPL-X kinematic chain with WiLoR's
          global wrist orientation via swing-twist decomposition, then
          refine only the shoulder using confidence-weighted MediaPipe
          keypoints, resolving shoulder-forearm inconsistencies without
          disrupting hand accuracy.
          
    \item \textbf{Generalization study.} The closed-form geometric design
          generalizes across five typologically distinct sign languages
          without dataset-specific adaptation, achieving up to 32\% lower
          hand error than prior methods at $32\times$ the speed of the
          strongest baseline.
\end{itemize}

\section{Related Work}

\subsection{Whole-Body 3D Pose Estimation}

Recent advances enable detailed 3D reconstruction from monocular images
using parametric body models. The foundational SMPL
model~\cite{SMPL:2015} represents the human body through shape and pose
parameters learned from diverse 3D meshes, providing a unified parametric
representation. SMPL-X~\cite{pavlakos2019expressivebodycapture3d} extends
this to incorporate expressive hands and face, enabling whole-body pose
estimation with a unified parameter space.

Regression-based methods directly predict body parameters from images.
Kanazawa et al.~\cite{kanazawaHMR18} introduced HMR, which uses
adversarial training to recover 3D meshes without paired 3D supervision;
SPIN~\cite{kolotouros2019learningreconstruct3dhuman} closed the loop by
incorporating SMPLify optimization within the training loop.
FrankMocap~\cite{rong2020frankmocapfastmonocular3d} employs separate
modules for body, hands, and face but achieves limited integration.
PIXIE~\cite{feng2021collaborativeregressionexpressivebodies} uses
collaborative regression for expressive bodies, and
PyMAF-X~\cite{zhang2023pymafx} refines whole-body fits via
mesh-aligned feedback. Video-based methods such as
VIBE~\cite{kocabas2020vibe} exploit temporal context but operate
on the body only, omitting expressive hand and face detail.
Hybrid regression-IK approaches such as
HybrIK~\cite{li2021hybrik} combine learned pose estimation with
analytic inverse kinematics, a design tradition our forearm
alignment builds on.
OSX~\cite{lin2023onestage3dwholebodymesh} introduces a one-stage
transformer with component-aware attention, and
SMPLer-X~\cite{cai2024smplerxscalingexpressivehuman} scales up with
a ViT-Huge backbone, demonstrating state-of-the-art performance
across multiple benchmarks. 2D pose detectors such as
MediaPipe~\cite{lugaresi2019mediapipe} complement these methods by
providing real-time confidence-weighted keypoint supervision that resolves
depth ambiguities in shoulder and elbow joints, a role our pipeline
explicitly exploits.

\subsection{Hand Pose Estimation}

Accurate 3D hand pose estimation is challenging due to self-occlusions,
complex articulations, and depth
ambiguity~\cite{zheng2021deep,moon2022accurate3dhandpose}.
Model-based methods use MANO parametric hand
models~\cite{Romero_2017} to ensure anatomically plausible predictions.
Hand4Whole~\cite{moon2022accurate3dhandpose} focuses on hand pose within
whole-body estimation but treats hands independently from body context.
Transformer- and graph-based mesh reconstruction
(METRO~\cite{lin2021metro},
MeshGraphormer~\cite{lin2021meshgraphormer}) recover vertex positions
directly without an intermediate parametric stage.
HaMeR~\cite{pavlakos2024reconstructing} employs a transformer-based
architecture with ViT-Huge backbone, achieving strong results across
occlusions, hand-object interactions, and diverse viewpoints.
WiLoR~\cite{potamias2025wilorendtoend3dhand} introduces end-to-end hand
localization and reconstruction through transformer-based refinement,
achieving superior accuracy via automatic hand detection. Our pipeline
leverages WiLoR's precise hand predictions while addressing the
non-trivial challenge of integrating MANO-format outputs with SMPL-X
body parameters. Progress in this domain has been driven by
large-scale benchmarks including FreiHAND~\cite{Freihand2019},
InterHand2.6M~\cite{Moon_2020_ECCV_InterHand}, and
HO-3D~\cite{hampali2020honnotatemethod3dannotation}.

\subsection{Sign Language Avatar Reconstruction}

Sign language reconstruction methods have advanced substantially on
Western sign languages. SGNify~\cite{Forte23-CVPR-SGNify} introduces
linguistic priors for isolated signs using sign-language-aware
optimization constraints, though at high computational cost. Neural Sign
Actors~\cite{baltatzis2024neuralsignactorsdiffusion} provides the first
high-quality 3D SMPL-X annotations for the How2Sign
dataset~\cite{duarte2021how2signlargescalemultimodaldataset},
establishing critical infrastructure for ASL avatar-based applications,
an analogous contribution to what we provide for Arabic Sign Language.
DexAvatar~\cite{kundu2025dexavatar} proposes sign-language-aware priors
for optimization-based reconstruction, achieving 30.13mm body and 13mm
hand errors, but requires 21.60 seconds per frame and is evaluated
exclusively on Western sign languages.

The downstream value of accurate parametric annotations extends beyond
reconstruction. Sign-language production
systems~\cite{baltatzis2024neuralsignactorsdiffusion,saunders2020progressive}
and self-supervised sign-recognition
models~\cite{hu2023signbertplus} rely directly on annotation quality
, noisy or geometrically inconsistent SMPL-X parameters in training
data propagate errors into generated or recognised signing motions.
This makes the absence of 3D ArSL annotations not merely an
evaluation gap, but a barrier to the entire research ecosystem. As a
concrete demonstration of this downstream value, we use the released
Tamaththul3D annotations to train a gloss-conditioned 3D sign-language
production model for ArSL; its motion tokenizer and qualitative results
are detailed in the supplementary material.

\textbf{Arabic Sign Language Gap.}
Limited work addresses ArSL reconstruction.
ArabSign~\cite{arabsign} provides a continuous ArSL dataset with 9,335
samples, while KArSL~\cite{karsl} offers 502 isolated ArSL signs with
75,300 total samples. Large-scale recognition benchmarks including
WLASL~\cite{li2020word} and PHOENIX-2014~\cite{KOLLER2015108} have driven
progress in Western sign languages, but no prior work provides 3D
parametric annotations or specialized reconstruction methods for ArSL
avatar generation. All existing 3D reconstruction methods focus on ASL,
German SL, or British SL, with no consideration for ArSL's unique
characteristics.

\section{Method}

\subsection{Overview}

\begin{figure}[t]
    \centering
    \includegraphics[width=1\textwidth]{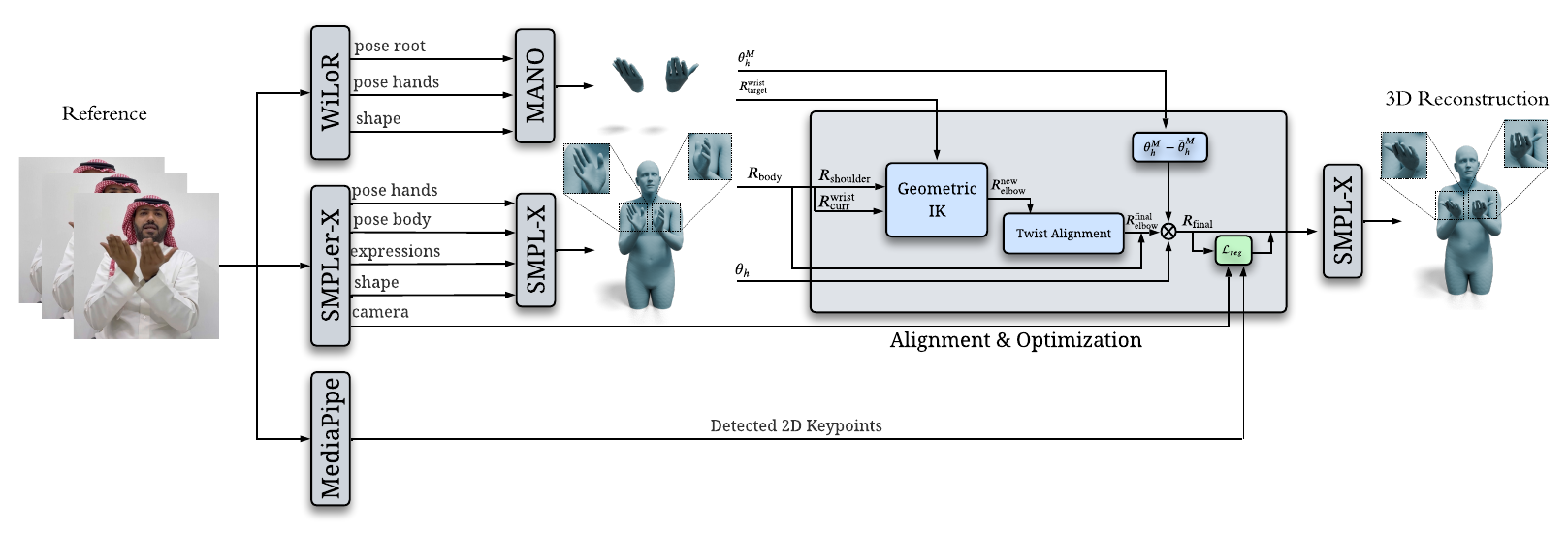}
    \caption{Tamaththul3D pipeline. SMPLer-X, WiLoR, and MediaPipe
        feed a geometric forearm-alignment stage and a 2D-supervised
        shoulder optimization, producing the final SMPL-X parameters.}
    \label{fig:pipeline}
\end{figure}

Tamaththul3D reconstructs expressive 3D avatars from monocular RGB video
by decoupling hand and body estimation and fusing them through geometric
inverse kinematics. The pipeline operates in four stages, illustrated in
\cref{fig:pipeline}. First, three complementary modules extract features
from each frame: SMPLer-X~\cite{cai2024smplerxscalingexpressivehuman}
estimates whole-body SMPL-X parameters, WiLoR~\cite{potamias2025wilorendtoend3dhand}
reconstructs detailed hand poses in MANO format, and
MediaPipe~\cite{lugaresi2019mediapipe} extracts confidence-weighted 2D
keypoints for supervision. Second, the geometric IK solver aligns the
forearm kinematic chain with WiLoR's accurate global wrist orientation.
Third, the aligned hand and body parameters are fused through coordinate
conversion, left-hand mirroring, and 2D-supervised shoulder optimization.
Finally, the corrected SMPL-X parameters render the output avatar.
Formally, the pipeline takes monocular RGB frames $\mathbf{I} \in
    \mathbb{R}^{H \times W \times 3}$ and outputs SMPL-X parameters
$\Theta = \{\beta, \theta_{\text{body}}, \theta_{\text{hand}}, \psi,
    \phi\}$ representing shape, body pose, hand pose, expression, and global
orientation respectively.

\subsection{Initial Pose Estimation}

Three off-the-shelf modules supply the inputs to our integration stage.
SMPLer-X~\cite{cai2024smplerxscalingexpressivehuman} regresses initial
whole-body SMPL-X parameters
$\Theta_{\text{init}} = \{\beta_{\text{init}},
    \theta_{\text{body}}^{\text{init}}, \psi_{\text{init}},
    \phi_{\text{init}}\}$.
WiLoR~\cite{potamias2025wilorendtoend3dhand} reconstructs each detected
hand $\mathbf{H}_i$ ($i \in \{\text{left}, \text{right}\}$) in MANO
format~\cite{Romero_2017}, producing finger joint rotations
$\theta_{\text{hand}}^i \in \mathbb{R}^{15 \times 3}$, hand shape
$\beta_{\text{hand}}^i \in \mathbb{R}^{10}$, global wrist rotation
$R_{\text{wrist}}^i \in SO(3)$, and translation
$t_i \in \mathbb{R}^3$. MediaPipe~\cite{lugaresi2019mediapipe} extracts
2D keypoints $\{\mathbf{k}_j, c_j\}$ with confidence $c_j \in [0,1]$ at
the shoulders, elbows, and wrists; high-confidence keypoints supervise
the optimization in \cref{sec:opt} while low-confidence ones are
down-weighted.

\subsection{Hand-Body Integration}
\label{sec:integration}

This is the central contribution of the pipeline (Stage~2 of
\cref{fig:pipeline}): integrating MANO-format hand poses from WiLoR
with SMPL-X body parameters from SMPLer-X. Direct substitution fails
due to: (1) different coordinate systems (MANO uses hand-centric
coordinates while SMPL-X uses body-centric; (2) different
parametrizations (MANO encodes poses relative to its own mean hand
pose, while SMPL-X uses a body-centric rest pose space; and (3) wrist
rotation ambiguity.

\textbf{Coordinate Conversion.}
To convert WiLoR's MANO poses to SMPL-X format, we convert rotation
matrices to axis-angle representation and subtract the MANO mean pose
$\bar{\theta}_{\text{hand}}^{\text{MANO}}$:
\begin{equation}
    \theta_{\text{hand}}^{\text{SMPL-X}} = \theta_{\text{hand}}^{\text{MANO}}
    - \bar{\theta}_{\text{hand}}^{\text{MANO}}
\end{equation}
This removes MANO's pose bias, yielding hand poses in SMPL-X's rest pose
space.

\textbf{Left Hand Mirroring.}
WiLoR processes left hands as mirrored right hands for model efficiency.
We apply a YZ-plane reflection to recover proper MANO left hand format:
\begin{equation}
    R_{\text{left}} = \mathbf{M} \cdot R_{\text{WiLoR}} \cdot
    \mathbf{M}^{\top}
\end{equation}
applied to both wrist rotation and all 15 finger joint rotations.

\textbf{Geometric Forearm Alignment.}
WiLoR provides highly accurate global wrist rotations in world space,
while SMPL-X requires local joint rotations in its kinematic tree. We
solve for the elbow rotation geometrically to ensure the entire forearm
chain matches WiLoR's wrist placement. Building the forward kinematics
chain:
\begin{equation}
    R_{\text{world}}^j = \begin{cases}
        R_{\text{local}}^0                               & \text{if } j = 0 \\
        R_{\text{world}}^{p(j)} \cdot R_{\text{local}}^j & \text{otherwise}
    \end{cases}
\end{equation}
where $p(j)$ is the parent of joint $j$. Given the target global wrist
rotation $R_{\text{target}}^{\text{wrist}}$ from WiLoR, we solve for
the elbow rotation that achieves exact alignment:
\begin{equation}
    R_{\text{elbow}}^{\text{local,new}} =
    (R_{\text{shoulder}}^{\text{world}})^{\top} \cdot
    R_{\text{target}}^{\text{wrist}} \cdot
    (R_{\text{wrist}}^{\text{local,cur}})^{\top}
\end{equation}

\textbf{Twist Extraction for Forearm Rotation.}
Forearm rotation (twist along the arm axis) requires special treatment.
We apply swing-twist
decomposition~\cite{dobrowolski2015swingtwistdecompositioncliffordalgebra}
to extract the twist component:
\begin{equation}
    \mathbf{a}_{\text{twist}} = \mathbf{f} \cdot (\mathbf{a}_{\text{rel}}
    \cdot \mathbf{f}), \quad \mathbf{a}_{\text{swing}} =
    \mathbf{a}_{\text{rel}} - \mathbf{a}_{\text{twist}}
\end{equation}
where $\mathbf{f}$ is the forearm axis and $\mathbf{a}_{\text{rel}}$ is
the relative rotation between target and current wrist configurations.
The twist is then applied to the geometric elbow solution:
\begin{equation}
    R_{\text{elbow}}^{\text{final}} = \exp(\mathbf{a}_{\text{twist}})
    \cdot R_{\text{elbow}}^{\text{local,new}}
\end{equation}

\subsection{2D-Supervised Upper Body Optimization}
\label{sec:opt}

After geometric forearm alignment (Stage~3 of \cref{fig:pipeline}),
the elbow and wrist joints are precisely positioned to match WiLoR's
hand predictions. However, the shoulder may require adjustment since
the geometric solution modifies the forearm without considering the
full arm chain. We optimize \textit{only the shoulder}
$\theta_{\text{shoulder}}$ while keeping the geometrically-aligned
elbow and wrist fixed, using a pose consistency loss:
\begin{equation}
    \mathcal{L} = \lambda_{\text{reg}} \left\lVert \theta_{\text{shoulder}}
    - \theta_{\text{shoulder}}^{\text{init}} \right\rVert^2
    + \lambda_{\text{2D}} \sum_{j \in J} w_j\, c_j \cdot
    \left\lVert \pi(\mathbf{p}_j) - \mathbf{k}_j \right\rVert_1
    + \lambda_{\text{pose}} \left\lVert \mathbf{z} \right\rVert^2
\end{equation}
where $\pi(\mathbf{p}_j)$ projects 3D joint positions to 2D and
$\mathbf{k}_j$ are MediaPipe keypoints with per-joint weights $w_j$
and confidence $c_j$. The final term is a learned body-pose prior
$\|\mathbf{z}\|^2$ on the
VPoser~\cite{pavlakos2019expressivebodycapture3d} latent encoding of
the full body pose, pushing the optimized shoulder toward
statistically plausible configurations; the supplementary material
discusses this prior in depth. Loss weights are optimized using Adam
with learning rate $1 \times 10^{-2}$ for 50 iterations per frame.



\subsection{Temporal Smoothing}

For video sequences, we suppress per-frame jitter through post-hoc
multi-order derivative minimization~\cite{smooth}:
\begin{equation}
    \mathcal{L}_{\text{temp}} = \lambda_{\text{data}}
    \mathcal{L}_{\text{data}} + \lambda_1 \mathcal{L}_{\text{d1}}
    + \lambda_2 \mathcal{L}_{\text{d2}} + \lambda_3
    \mathcal{L}_{\text{d3}},
\end{equation}
where the four terms enforce fidelity to the per-frame estimates and
penalize velocity, acceleration, and jerk respectively. Derivative
penalties are weighted more strongly for hand joints than body joints,
preserving fine finger articulation while suppressing torso and arm
noise. Side-by-side comparisons of smoothed vs.\ unsmoothed output
across multiple datasets are demonstrated qualitatively in
supplementary videos accompanying the annotation release.

\subsection{Modularity}
\label{sec:modularity}

The geometric integration in \cref{sec:integration} is closed-form and
makes no assumption about how its inputs were produced: it requires
only \emph{any} estimator that outputs SMPL-X body parameters and
\emph{any} estimator that outputs MANO-format hand parameters with a
global wrist rotation. SMPLer-X and WiLoR are selected as the current
best-performing choices, but neither the IK solver nor the
optimization stage depends on their internals, so improved body or
hand estimators can be substituted without re-deriving the alignment.
This decoupling distinguishes our approach from end-to-end methods
whose architectural assumptions are baked into the model weights, and
is validated empirically in \cref{sec:modularity_results} by swapping
the body and hand estimators independently.

\section{Experiments}

\subsection{Datasets}

\subsubsection{Ishara-500}
The Ishara-500 subset of the Ishara
dataset~\cite{alyami2025isharah} is a large-scale continuous Saudi
Sign Language dataset comprising 30,000 video samples captured in
unconstrained environments using smartphone cameras. Ishara-500
contains videos from 18 diverse signers performing over 500 unique
SSL sentences, with natural variations in camera angles, distances,
lighting conditions, and backgrounds.

\subsubsection{Ishara-500 Annotations}
This work produces the first high-quality SMPL-X parameter
annotations for Ishara-500, establishing it as the first 3D Arabic
Sign Language dataset with parametric avatar representations.
Annotations will be publicly released to enable future research in
Arabic Sign Language avatar reconstruction and related applications.
The dataset was collected and released by Alyami et
al.~\cite{alyami2025isharah} under their institutional review board
(IRB) approval; this work performs no new human-subject data
collection and uses Ishara-500 solely for annotation and evaluation
purposes, so no additional IRB review was required.

\subsubsection{Cultural Clothing and Model Bias}
\label{sec:clothing}
A significant challenge when processing Ishara-500 concerns
traditional Saudi attire: male signers frequently wear thobes
(ankle-length white robes) and female signers wear hijabs and
abayas. These culturally significant garments are absent from the
training data of foundation models like SMPLer-X, which were trained
predominantly on Western datasets with form-fitting clothing. The
loose, flowing fabric obscures body shape cues and joint locations,
presenting a critical case of model bias when applying
state-of-the-art methods to underrepresented populations.
MediaPipe-guided 2D optimization partially addresses this by
providing clothing-invariant keypoint guidance for shoulder and
elbow refinement.

\subsubsection{SGNify Benchmark}
For quantitative evaluation and comparison with prior methods, the
SGNify mocap dataset~\cite{Forte23-CVPR-SGNify} is used, containing
ground-truth SMPL-X annotations for sign language sequences captured
with high-quality motion capture, enabling direct numerical
comparison with existing reconstruction methods.

\subsection{Evaluation Metrics}

Procrustes-Aligned Mean Per Vertex Position Error
(PA-MPVPE)~\cite{pavlakos2019expressivebodycapture3d,
    kanazawaHMR18} is reported in millimeters, computed separately for
body, left hand, and right hand regions. PA-MPVPE applies Procrustes
alignment~\cite{Gower1975} before computing distances,
making the metric invariant to global rotation and translation while
focusing on shape and pose accuracy.

\subsection{Implementation Details}

\textbf{Components.}
Official pretrained weights for
SMPLer-X~\cite{cai2024smplerxscalingexpressivehuman} and
WiLoR~\cite{potamias2025wilorendtoend3dhand} are used without any
fine-tuning, demonstrating strong out-of-the-box generalization.
MediaPipe~\cite{lugaresi2019mediapipe} uses default Pose and Hand
models with no modifications.

\textbf{Preprocessing.}
Input frames are processed at their original resolution. Person
bounding boxes are detected via the integrated detector in SMPLer-X.
Hand regions are automatically localized by WiLoR's built-in
detection module, removing the need for manual cropping or region
proposals.

\textbf{SMPL-X and MANO Models.}
The neutral-gender SMPL-X
model~\cite{pavlakos2019expressivebodycapture3d} is used with 10
shape coefficients and 10 expression coefficients, without PCA
compression, to preserve the full articulation range. MANO
models~\cite{Romero_2017} use 45 PCA components with the flat hand
mean disabled to ensure compatibility with SMPL-X's hand pose space.

\textbf{Camera Model.}
Camera intrinsics (focal length and principal point) are extracted
directly from SMPLer-X predictions. A weak-perspective camera model
is used for all 2D-to-3D correspondences during optimization.

\textbf{Runtime.}
All experiments were conducted on an NVIDIA RTX 5070 Ti with 16GB
GPU memory and 32GB CPU memory. The full pipeline processes a
150-frame SSL video at 30 fps in roughly 100 seconds (0.67
s/frame), compared to 54 minutes (21.60 s/frame) for DexAvatar on
identical hardware.

\subsection{Results}

\subsubsection{Main Results}

Tamaththul3D is benchmarked against seven existing methods on the
SGNify dataset, spanning general whole-body estimators,
sign-language-specific reconstruction methods, and the strongest
prior baseline DexAvatar. \cref{tab:main_results} reports PA-MPVPE
for body, left hand, and right hand; per-frame runtime against the
strongest baseline is reported separately in \cref{tab:mpja}.

\begin{table}
    \centering
    \caption{Quantitative comparison on the SGNify dataset.
        PA-MPVPE (mm) measures geometric accuracy; lower is better.}
    \label{tab:main_results}
    \small
    \setlength{\tabcolsep}{6pt}
    \begin{tabular}{lccc}
        \toprule
        \textbf{Method}                                              & \textbf{Body $\downarrow$} & \textbf{L.~Hand $\downarrow$} & \textbf{R.~Hand $\downarrow$} \\
        \midrule
        FrankMocap~\cite{rong2020frankmocapfastmonocular3d}          & 78.07                      & 20.47                         & 19.62                         \\
        PIXIE~\cite{feng2021collaborativeregressionexpressivebodies} & 60.11                      & 25.02                         & 22.42                         \\
        SMPLify-X~\cite{pavlakos2019expressivebodycapture3d}         & 56.07                      & 22.23                         & 18.83                         \\
        SGNify~\cite{Forte23-CVPR-SGNify}                            & 55.63                      & 19.22                         & 17.50                         \\
        OSX~\cite{lin2023onestage3dwholebodymesh}                    & 60.79                      & 19.10                         & 18.79                         \\
        NSA~\cite{baltatzis2024neuralsignactorsdiffusion}            & 46.42                      & 16.17                         & 15.23                         \\
        DexAvatar~\cite{kundu2025dexavatar}                          & 30.13                      & 13.53                         & 13.08                         \\
        \midrule
        \textbf{Tamaththul3D}                                        & \textbf{29.28}             & \textbf{10.65}                & \textbf{8.90}                 \\
        \bottomrule
    \end{tabular}
\end{table}

Tamaththul3D achieves state-of-the-art hand accuracy across all
baselines while running $32\times$ faster than the strongest baseline
DexAvatar (\cref{tab:mpja}). General methods (FrankMocap, PIXIE,
SMPLify-X) produce poor hand accuracy (18--25mm) as they are not
designed for fine-grained hand articulation. Sign-specific methods
(SGNify, NSA) improve hands (15--19mm) but remain insufficient for
semantic clarity. The NSA~\cite{baltatzis2024neuralsignactorsdiffusion}
fitting pipeline was confirmed unavailable for external release at
the time of submission following direct correspondence with the
authors.

\subsubsection{Ablation Study}


\begin{table}[h!]
    \centering
    \caption{Ablation study on SGNify dataset (PA-MPVPE in mm).}
    \label{tab:ablation}
    \small
    \setlength{\tabcolsep}{6pt}
    \begin{tabular}{lccc}
        \toprule
        \textbf{Configuration}                        & \textbf{Body $\downarrow$} & \textbf{L.~Hand $\downarrow$} & \textbf{R.~Hand $\downarrow$} \\
        \midrule
        SMPLer-X                                      & 28.46                      & 18.17                         & 17.47                         \\
        W/ 2D Supervision                             & \textbf{28.35}             & 18.17                         & 17.47                         \\
        W/ WiLoR (\textbf{C}oord.\ \textbf{C}onv.)    & 28.46                      & 10.71                         & 9.03                          \\
        W/ WiLoR (\textbf{G}eometric \textbf{A}lign.) & 29.53                      & 10.68                         & 8.95                          \\
        \midrule
        \textbf{Full pipeline}                        & 29.28                      & \textbf{10.65}                & \textbf{8.90}                 \\
        \bottomrule
    \end{tabular}
\end{table}

\begin{wrapfigure}[16]{r}{0.4\linewidth}
    \centering
    \vspace{-4mm}
    \includegraphics[width=\linewidth,
        keepaspectratio]{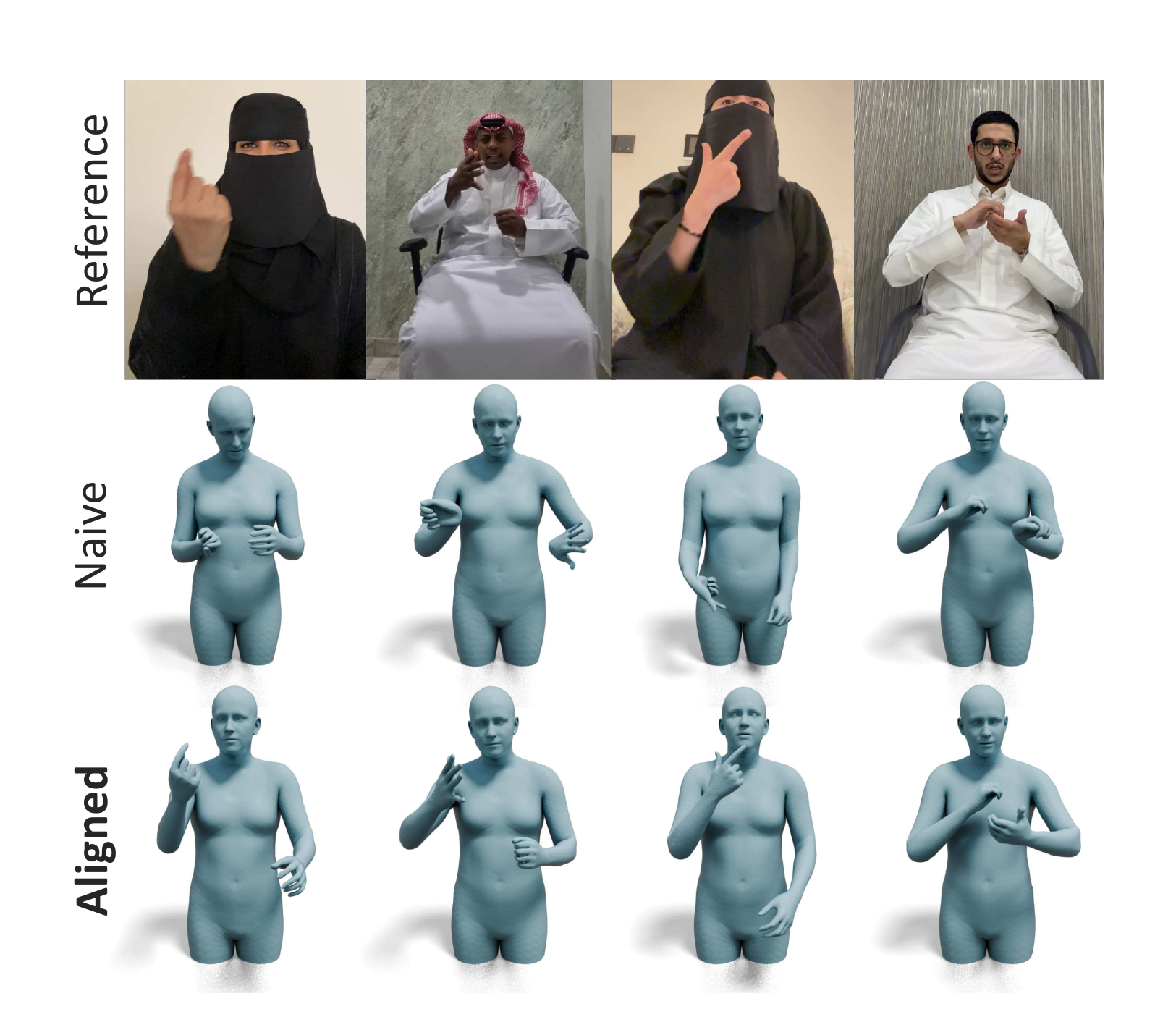}
    \caption{\textbf{Kinematic artifacts} without geometric forearm
        alignment. Top: reference; middle: naive WiLoR substitution;
        bottom: with geometric alignment.}
    \label{fig:kinematic_artifacts}
\end{wrapfigure}
Each stage of the pipeline contributes incrementally to the final
result. SMPLer-X alone produces robust body pose estimates (28.46mm)
but poor hand accuracy (18.17mm left, 17.47mm right), insufficient
for sign language semantics. Adding 2D supervision alone yields only
marginal improvement on the body (28.35mm) and none on the hands,
confirming that 2D keypoints without explicit hand refinement have
limited effect.  Coordinate conversion replaces SMPL-X hand poses with
WiLoR predictions, producing a large improvement in hand accuracy
(10.71mm left, 9.03mm right, a 48\% reduction in right-hand error)
while preserving body pose (28.46mm). Geometric alignment further
refines results by solving for elbow rotation aligned with WiLoR's
global wrist (8.95mm right). Geometric alignment yields only marginal
numerical improvement over coordinate substitution, but its primary
value is qualitative: it guarantees kinematic consistency between
body and hand, preventing ``broken wrist'' artifacts that occur with
naive substitution (\cref{fig:kinematic_artifacts}). The full pipeline
achieves the best overall trade-off, with competitive body pose
(29.28mm, a 0.82mm degradation from baseline, $<3\%$) and the most
accurate hands (10.65mm left, 8.90mm right).

\subsubsection{Modularity Analysis}
\label{sec:modularity_results}

To empirically validate the modularity claim of \cref{sec:modularity},
we replace the body and hand estimators independently and re-evaluate
on SGNify. We consider three body backbones (OSX~\cite{lin2023onestage3dwholebodymesh},
SMPLest-X~\cite{yin2025smplest}, and
SMPLer-X~\cite{cai2024smplerxscalingexpressivehuman}) and three hand
estimators (HaMeR~\cite{pavlakos2024reconstructing},
Hamba~\cite{dong2024hamba}, and
WiLoR~\cite{potamias2025wilorendtoend3dhand}), yielding nine
combinations plus body-only baselines (\cref{tab:modularity_ablation}).

\begin{table}[t]
    \centering
    \caption{Modularity ablation on the SGNify
        dataset~\cite{Forte23-CVPR-SGNify}. PA-MPVPE (mm) reported per
        region; lower is better. \textbf{Bold} indicates the best
        result per column overall; \underline{underline} indicates the
        best result per column among configurations that include a
        hand module.}
    \label{tab:modularity_ablation}
    \small
    \setlength{\tabcolsep}{6pt}
    \begin{tabular}{llccc}
        \toprule
        \textbf{Body} & \textbf{Hands}                               & \textbf{Body} $\downarrow$ & \textbf{L.~Hand} $\downarrow$ & \textbf{R.~Hand} $\downarrow$ \\
        \midrule
        \multirow{4}{*}{OSX~\cite{lin2023onestage3dwholebodymesh}}
                      & \textit{None}                                & \underline{60.79}          & 19.10                         & 18.79                         \\
                      & HaMeR~\cite{pavlakos2024reconstructing}      & 61.06                      & 12.36                         & 9.38                          \\
                      & Hamba~\cite{dong2024hamba}                   & 62.20                      & 12.47                         & 9.58                          \\
                      & WiLoR~\cite{potamias2025wilorendtoend3dhand} & 61.69                      & \underline{12.03}             & \underline{8.92}              \\
        \midrule
        \multirow{4}{*}{SMPLest-X~\cite{yin2025smplest}}
                      & \textit{None}                                & \underline{35.68}          & 16.64                         & 18.37                         \\
                      & HaMeR~\cite{pavlakos2024reconstructing}      & 37.14                      & 11.84                         & 9.42                          \\
                      & Hamba~\cite{dong2024hamba}                   & 37.11                      & 12.13                         & 9.69                          \\
                      & WiLoR~\cite{potamias2025wilorendtoend3dhand} & 36.55                      & \underline{11.47}             & \underline{8.94}              \\
        \midrule
        \multirow{4}{*}{SMPLer-X~\cite{cai2024smplerxscalingexpressivehuman}}
                      & \textit{None}                                & \underline{\textbf{28.46}} & 18.17                         & 17.47                         \\
                      & HaMeR~\cite{pavlakos2024reconstructing}      & 30.07                      & 12.30                         & 9.36                          \\
                      & Hamba~\cite{dong2024hamba}                   & 30.12                      & 11.54                         & 9.65                          \\
                      & WiLoR~\cite{potamias2025wilorendtoend3dhand} & 29.28                      & \underline{\textbf{10.65}}    & \underline{\textbf{8.90}}     \\
        \bottomrule
    \end{tabular}
\end{table}

\textbf{Findings.}
\Cref{tab:modularity_ablation} supports three claims about the
modularity of the design.
(1)~\textit{Hand accuracy is invariant to body-backbone choice.}
Body PA-MPVPE varies by over 30~mm across the three body estimators
(OSX 60.79, SMPLest-X 35.68, SMPLer-X 28.46), yet right-hand error
varies by at most $0.11$~mm across body backbones for any fixed hand
module (\eg WiLoR yields 8.90/8.92/8.94 across the three bodies).
The geometric IK stage therefore propagates hand accuracy
independently of the body backbone.
(2)~\textit{Hand-module ranking is consistent across body backbones.}
For every body backbone, right-hand error obeys WiLoR $<$ HaMeR $<$
Hamba: the integration composes predictably with the hand estimator
and scales with its accuracy rather than being tuned to any specific
model.
(3)~\textit{The 2D-supervised shoulder refinement preserves body
    fidelity.} Adding any hand module degrades body PA-MPVPE by at most
$\sim$2~mm, confirming that the optimization does not disrupt the
body pose produced by the chosen backbone.
Together these results show that the geometric IK integration is
decoupled from the specific choice of body and hand estimators, and
that SMPLer-X + WiLoR is the strongest combination among current
estimators rather than a configuration the pipeline is tied to.



\subsubsection{Temporal Smoothness}

As quantified in \cref{tab:mpja}, temporal smoothing produces
substantially more stable reconstructions than the per-frame baseline.
Following WiLoR~\cite{potamias2025wilorendtoend3dhand}, we report
\textit{Jitter} (the mean magnitude of the third derivative
(jerk) of joint position, which isolates high-frequency motion
noise from intentional articulation), and \textit{RTE}, the mean
frame-to-frame wrist displacement, which captures spatial
stability. Compared to DexAvatar's per-frame output, Tamaththul3D
reduces hand jitter by 83.2\%, body jitter by 83.9\%, and RTE by
62.3\%. The evaluation spans a 560-frame multi-signer sequence with
finite-difference windows masked at signer boundaries so scene cuts
do not inflate the metrics.

\begin{figure}[t]
    \centering
    \begin{minipage}[b]{0.49\textwidth}
        \centering
        \includegraphics[width=\linewidth]{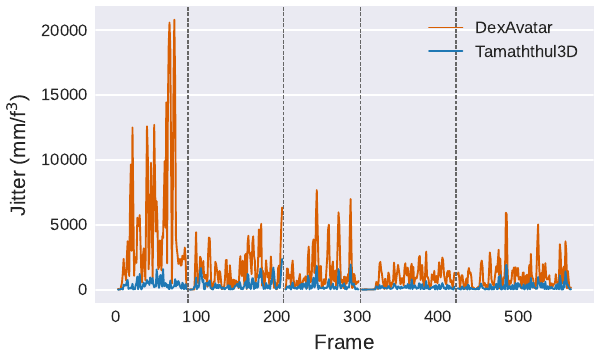}
        \captionof{figure}{Per-frame upper-body + hand jitter (mm/f$^3$).
            Dashed lines mark signer boundaries.}
        \label{fig:jitter_time}
    \end{minipage}\hfill
    \begin{minipage}[b]{0.49\textwidth}
        \centering
        \includegraphics[width=\linewidth]{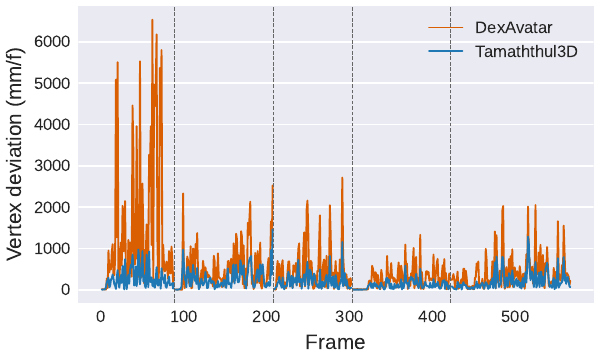}
        \captionof{figure}{Per-frame mesh vertex deviation (mm/f),
            capturing total avatar surface motion frame-to-frame.}
        \label{fig:vertex_dev}
    \end{minipage}
\end{figure}

\begin{table}[t]
    \centering
    \caption{Runtime and temporal stability on 560 frames spanning
        five signers (identical hardware). Jitter and RTE follow
        WiLoR~\cite{potamias2025wilorendtoend3dhand}. Lower is better.}
    \label{tab:mpja}
    \small
    \setlength{\tabcolsep}{6pt}
    \begin{tabular}{lcccc}
        \toprule
        \textbf{Method}                     & \textbf{Time (s/f) $\downarrow$} & \textbf{Jitter Hands $\downarrow$} & \textbf{Jitter Body $\downarrow$} & \textbf{RTE $\downarrow$} \\
        \midrule
        DexAvatar~\cite{kundu2025dexavatar} & 21.60                            & 1783.64                            & 1791.15                           & 572.52                    \\
        \midrule
        \textbf{Tamaththul3D}               & \textbf{0.67}                    & \textbf{299.14}                    & \textbf{289.02}                   & \textbf{215.53}           \\
        \bottomrule
    \end{tabular}
\end{table}

\Cref{fig:jitter_time,fig:vertex_dev} visualise this stability
frame-by-frame. DexAvatar (orange) exhibits sharp, high-amplitude
spikes throughout every signer segment: these are the per-frame
optimisation residuals that manifest as visible avatar jitter.
while Tamaththul3D (blue) tracks a substantially lower baseline,
with isolated low-amplitude excursions corresponding to genuine sign
articulation. The two metrics agree: joint-space jerk and full-mesh
vertex deviation both show the same DexAvatar–Tamaththul3D
contraction across all five signer segments, confirming that the
smoothing acts uniformly across body, hands, and the resulting mesh
surface. This stability is essential for the rendered avatar to be
intelligible to a viewer: a jittery output is perceived as noise
regardless of per-frame reconstruction accuracy.

\subsubsection{Generalization Across Sign Languages}

\begin{figure}
    \vspace{-6mm}
    \centering
    \includegraphics[width=\textwidth]{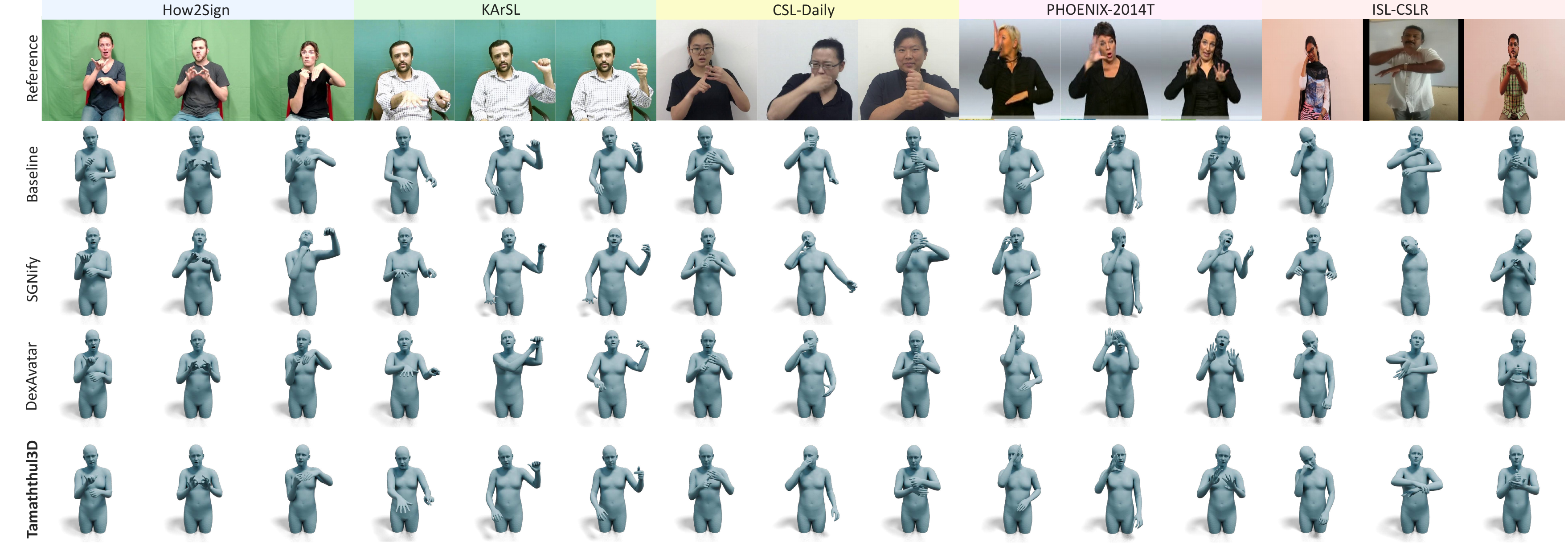}
    \caption{Qualitative generalization across five sign language
        datasets:
        How2Sign~\cite{duarte2021how2signlargescalemultimodaldataset},
        KArSL~\cite{karsl}, CSL-Daily~\cite{csl-daily},
        PHOENIX-2014T~\cite{KOLLER2015108}, and
        ISL-CSLTR~\cite{idiansl}. Rows: reference, SMPLer-X,
        SGNify~\cite{Forte23-CVPR-SGNify},
        DexAvatar~\cite{kundu2025dexavatar}, Tamaththul3D.}
    \label{fig:generalization}
\end{figure}

Tamaththul3D is a fully closed-form pipeline with no learned
parameters of its own, meaning its geometric IK solver and
coordinate conversion apply universally without dataset-specific
adaptation. The pipeline is applied to five sign language datasets
spanning four languages and three continents.
As shown in \cref{fig:generalization}, Tamaththul3D consistently
outperforms all baselines across all five datasets. SGNify, relying
on ASL/GSL linguistic priors, produces severe kinematic artifacts on
out-of-distribution signing styles. DexAvatar exhibits artifacts
particularly on KArSL and ISL-CSLTR, where learned optimization
priors generalize poorly outside Western sign language distributions.
The geometry-driven approach is unaffected by this distribution
shift, confirming that closed-form geometric alignment generalizes
where learned priors do not.

\subsubsection{Evaluation on Ishara-500}

\begin{figure}
    \centering
    \includegraphics[width=\textwidth]{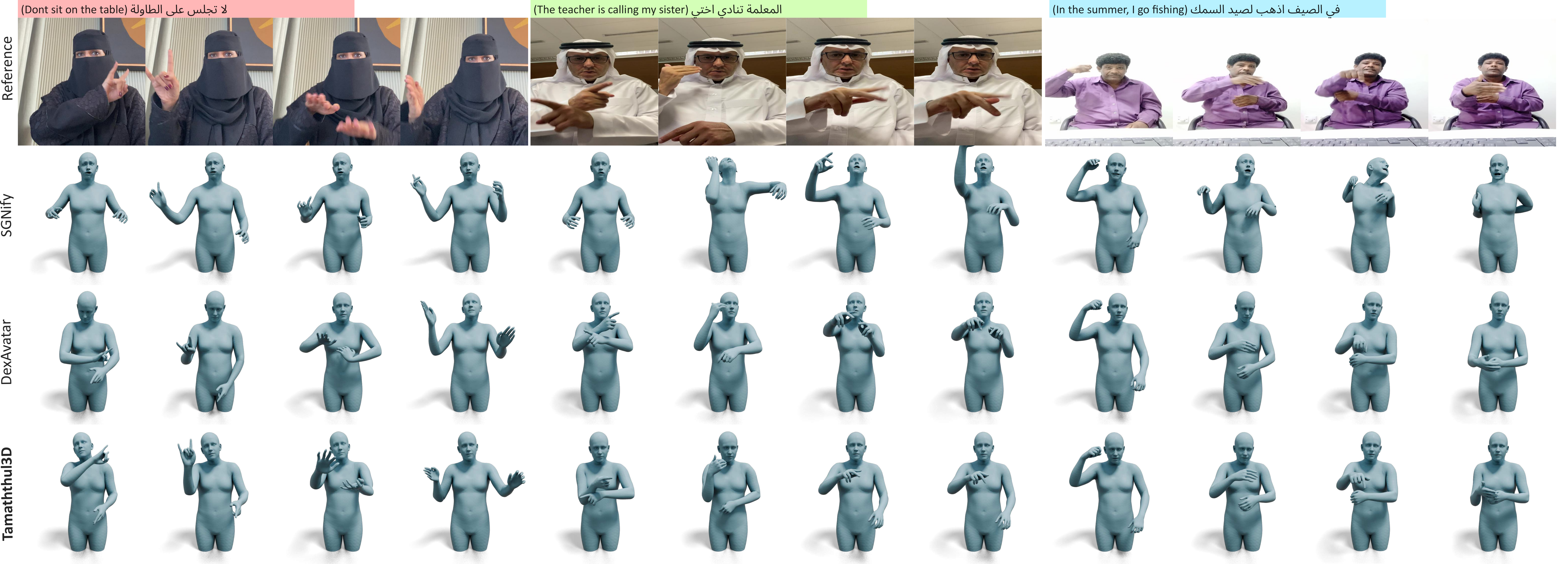}
    \caption{Qualitative evaluation on Ishara-500. Three signers in
        culturally distinct attire (abaya/niqab, thobe/guthra,
        form-fitting); four frames per sentence with translation
        above. Rows: reference, SGNify~\cite{Forte23-CVPR-SGNify},
        DexAvatar~\cite{kundu2025dexavatar}, Tamaththul3D (ours).}
    \label{fig:isharah_eval}
\end{figure}

\cref{fig:isharah_eval} evaluates Tamaththul3D on Ishara-500 across
three signers wearing culturally distinct attire. Since no 3D
ground-truth annotations exist for any Arabic Sign Language dataset,
evaluation is qualitative, a limitation of the field's current
annotation infrastructure that the released Ishara-500 annotations
are intended to help address. Compared to DexAvatar, Tamaththul3D
produces more accurate hand articulation and anatomically plausible
body posture across all three cultural clothing conditions.

\section{Limitations}

Tamaththul3D inherits constraints from each constituent component. 
The most persistent challenge is traditional Saudi clothing: SMPLer-X 
was trained predominantly on form-fitting Western attire, so thobes 
and abayas cause the model to lose the silhouette cues it relies on 
for shoulder and torso localization. MediaPipe 2D supervision 
partially compensates by providing clothing-invariant keypoint 
guidance, but underlying shape estimation remains unreliable in 
these cases.

WiLoR's hand localization fails predictably under two conditions:
severe inter-hand occlusion and extreme lateral viewing angles, which
together affect roughly 5\% of frames; fallback to SMPLer-X's native
hand estimates in these cases produces noticeably weaker results.

Finally, quantitative evaluation is constrained by the absence of 3D 
ground-truth annotations for any Arabic Sign Language dataset. The 
SGNify benchmark covers Western sign languages; SSL results are 
therefore evaluated perceptually rather than numerically, a 
limitation of the field's annotation infrastructure that the released 
Ishara-500 annotations are intended to help address.

\section{Conclusion}

We presented Tamaththul3D, a reconstruction pipeline for 3D sign
language avatars that integrates
SMPLer-X~\cite{cai2024smplerxscalingexpressivehuman},
WiLoR~\cite{potamias2025wilorendtoend3dhand}, and
MediaPipe~\cite{lugaresi2019mediapipe} through closed-form geometric
forearm alignment and 2D-supervised shoulder refinement. The
integration is decoupled from any specific body or hand estimator:
substituting alternative backbones preserves hand accuracy to within
$0.1$~mm, making the design directly reusable as the underlying
estimators improve. The pipeline achieves state-of-the-art hand
accuracy while maintaining competitive body pose, and generalizes without
dataset-specific adaptation. Applied to
Ishara-500~\cite{alyami2025isharah}, Tamaththul3D produces the first
SMPL-X annotations for any Arabic Sign Language dataset, providing
the foundation for downstream accessibility technologies in
education, telecommunication, and cultural preservation for the Arab
Deaf community.

\textbf{Ethical Statement.}
This work performs no new human-subject data collection.
Ishara-500~\cite{alyami2025isharah} and the SGNify
benchmark~\cite{Forte23-CVPR-SGNify} are used under their respective
institutional review board approvals, solely for annotation and
evaluation purposes; no additional IRB review was required.

%
%
\bibliographystyle{unsrt}
\bibliography{references}

\end{document}